\documentclass[10pt,twocolumn,letterpaper]{article}

\usepackage{wacv}              

\definecolor{wacvblue}{rgb}{0.21,0.49,0.74}
\usepackage[pagebackref,breaklinks,colorlinks,allcolors=wacvblue]{hyperref}
\usepackage[accsupp]{axessibility}
\def\wacvPaperID{8} 
\def\confName{WACV}
\def\confYear{2027}

\title{A Benchmark Protocol and Fairness Analysis Towards Open-Set EEG Seizure Type Classification}

\author{Targol Bakhtiarvand \quad Terrance E. Boult \quad Jugal Kalita \quad Adham Atyabi\\
University of Colorado Colorado Springs\\
{\tt\small \{tbakhtia, tboult, jkalita, aatyabi\}@uccs.edu}
}

\begin{document}
\maketitle
\begin{abstract}
Automated EEG seizure type classification systems are typically trained
and evaluated under the closed-set assumption, where all seizure types
encountered at deployment are seen during training. In clinical practice,
this assumption fails: patients present with rare or atypical seizure
types that fall outside the training label space, and a system that
forces a wrong label is worse than one that abstains. We introduce the
first open-set recognition protocol for EEG seizure type classification
on the Temple University Hospital Seizure Corpus (TUSZ v2.0.3), defining
known and unknown classes using the clinical taxonomy itself: seizure
types with sufficient subject representation ($\geq$12 subjects) form
the known set, and rare types form the unknown set. We evaluate seven post-hoc open-set scorers (MSP, MaxLogit, energy, Mahalanobis, kNN, GHOST, and COSTARR) applied to a partially fine-tuned EEG Foundation Model (EFM), using a subject-disjoint leave-one-class-out (LOCO) protocol over three seeds that tests whether methods can detect seizure types excluded from fine-tuning. On true unknown classes, the best scorer (Mahalanobis) reaches a mean AUROC of 0.66, and differences among the top scorers are within seed-to-seed variation. Held-out known seizure types from unseen subjects are detected only slightly above chance (AUROC 0.54–0.58 for all scorers). Per-subject rejection rates are highly uneven for all scorers, and GHOST does not significantly improve per-subject fairness over energy scoring (two-sided Wilcoxon signed-rank, $p \geq 0.25$). We release the protocol, subject splits, and code as an initial baseline; our results indicate that open-set seizure type recognition remains a hard, open problem.
\end{abstract}
\section{Introduction}
Epilepsy is a leading cause of neurological disability worldwide. Approximately 52 million people globally were affected by epilepsy in 2021, and there are many cases of disability and death \cite{feigin2025global}. Clinical seizure classification is important because epilepsy diagnosis and treatment vary depending on the type of seizure, epilepsy type, and syndrome. They do not simply vary because seizures occurred \cite{Kanner2022Antiseizure}. ILAE classification systems use seizure type as the foundational step in epilepsy diagnosis, and further classify seizures as focal, generalized, unknown, or unclassified because differences in seizure description have clinical management implications \cite{beniczky2025updated}.

Automated EEG systems have made steady progress in seizure detection and seizure-type classification, particularly on the Temple University Hospital Seizure Corpus (TUSZ), the main open public benchmark for multi-class seizure typing \cite{Rivera2024Evaluating}. Prior studies typically formulate the task as a closed-set K-class problem, training and evaluating models on a fixed set of seizure labels such as focal non-specific, generalized non-specific, absence, tonic, tonic-clonic, complex partial, simple partial, and sometimes myoclonic seizure \cite{abirami2024automated}. Representative approaches include feature-based machine learning, CNNs, CNN-LSTM hybrids, and transfer-learning pipelines, with reported performance improvements under standard inter-patient or cross-validation protocols \cite{Albaqami2022MP-SeizNet:, raghu2020eeg}. Reviews of this literature likewise organize the field around focal-versus-generalized or multi-class seizure classification within the annotated TUSZ label space, and identify the need for more standardized benchmarking rather than any open-set evaluation protocol \cite{McCallan2023Epileptic}.

That closed-set assumption poorly matches clinical deployment. TUSZ itself contains severe class imbalance and very limited representation for several seizure types, with some labels excluded entirely in prior work because too few patients or events were available for stable training \cite{abirami2024automated,akor2025hierarchical}. Several studies therefore restrict analysis to five or seven seizure classes, omit myoclonic seizures, or exclude non-specific labels to make the classification problem more tractable. Yet the non-specific labels in TUSZ are clinically meaningful, not mere annotation noise. When there is no clear evidence for a more specific label, events are assigned as focal non-specific or generalized non-specific. This means a deployed system will inevitably encounter ambiguous, atypical, or rare seizure presentations that do not fit neatly into the label set seen during fine-tuning.

Open-set recognition addresses exactly this setting by requiring a model to classify known classes while rejecting samples from unseen classes as unknown \cite{Sun2023Research}. The motivation is both practical and statistical: standard softmax classifiers are inherently closed-set and tend to produce overconfident errors when test inputs fall outside the training label space \cite{Xue2026OpenEEG:}. Recent OSR work in EEG and related signal domains shows growing recognition of this problem, including uncertainty-aware diagnosis frameworks for EEG brain disorders and prototype-, transformer-, or generative-based methods for open-set signal recognition \cite{Han2025Multi-layer}. However, these methods have not established a benchmark protocol for open-set seizure type recognition on TUSZ, and prior signal-domain OSR work explicitly notes that many existing OSR successes remain concentrated in image or text settings rather than clinically relevant signals \cite{Hou2025Multi-View}.

This gap motivates the present work. Rather than proposing a new seizure classifier under the standard closed-set assumption, we introduce the first open-set protocol for EEG seizure type classification on TUSZ and use it to baseline existing OSR techniques on a clinically meaningful task. Our protocol defines known and unknown seizure classes using the seizure taxonomy and the practical constraints imposed by class frequency, and it includes leave-one-class-out (LOCO) evaluation to test whether methods can detect seizure types excluded from fine-tuning. We evaluate seven post-hoc open-set scorers, including energy score~\cite{Liu2020Energy}, GHOST~\cite{ghost2025}, and COSTARR~\cite{costarr2025}, under a subject-disjoint partition with three seeds, and analyze both overall detection and the per-subject consistency of rejection. This framing is consistent with the clinical role of seizure taxonomies, which are intended to support communication and management under varying degrees of certainty rather than to force every event into an overly specific label.

We further study this problem on top of a pretrained EEG foundation model
rather than from scratch. Recent EEG representation learning work has
emphasized transferable backbones for downstream seizure tasks, and
multimodal pretraining has been shown to learn seizure-relevant EEG
representations that fine-tune effectively across benchmarks.
Building on that line of work, we evaluate whether a pretrained EEG
foundation model can provide embeddings suitable for open-set scoring
after parameter-efficient fine-tuning on known seizure types. In this
sense, the contribution is not only a new evaluation setting, but also
a first step toward clinically deployable seizure typing systems that
can recognize the limits of their label space instead of forcing
incorrect predictions.
\section{Related works}
Our work sits at the intersection of EEG representation learning, seizure type classification, and open-set recognition. We review each in turn and identify the gap our work addresses.
\subsection{EEG Foundation Models and Transfer Learning}
Recently proposed EEG foundation models aim to learn transferable representations from large amounts of heterogeneous unlabeled recordings instead of learning task-specific models from scratch \cite{Liu2026EEG-FM-Compass:, jiang2024large, chen2024eegformer}. LaBraM is a prominent example: it segments EEG into channel patches, pretrains Transformers with masked neural-code prediction, and is trained on about 2,500 hours from around 20 datasets before transfer to multiple downstream tasks \cite{jiang2024large}.

The architecture most closely related to our setting is our prior
multimodal EEG foundation model~\cite{bakhtiarvand2026efm}, which joins a raw-signal Mamba encoder, a ViT-style time--frequency encoder, and a lightweight text encoder in a joint embedding space. Its pretraining objective unifies masked modeling, cross-view contrastive alignment, and
temporal consistency losses to capture seizure-relevant structure without labels. Fine-tuned, it reaches state-of-the-art results on CHB-MIT~\cite{PhysioNet-chbmit-1.0.0}.  It also reports a leave-one-subject-out evaluation, highlighting that patient-independent seizure detection remains difficult despite strong standard-split results. Other recent pretraining frameworks, including DMAE-EEG and NeuroTTT, likewise emphasize cross-task transfer, denoising, calibration, or pretraining–downstream alignment rather than unknown-class handling at deployment \cite{Zhang2025DMAE-EEG:, wang2025neurottt}. Benchmarking studies further show that linear probing is often insufficient and that current EEG foundation models do not automatically yield robust real-world generalization \cite{Liu2026EEG-FM-Compass:}. However, none of these pretraining and transfer-learning works study the clinically important case where a fine-tuned seizure model encounters seizure types unseen during training \cite{bakhtiarvand2026efm}.
\subsection{Seizure Detection and Classification}
The seizure modeling literature, which our work builds upon, largely considers closed-set seizure detection or classification. Seizure versus non-seizure segment classification with deep architectures has mainly used CNNs, recurrent models, or hybrids thereof on CHB-MIT and TUSZ \cite{shah2018temple}, reporting high accuracy under conventional protocols \cite{Thuwajit2022EEGWaveNet:, Amiri2025Epileptic, Cao2025A}. EEGWaveNet, a multiscale end-to-end CNN, has been assessed on CHB-MIT and TUSZ, reporting improved subject-independent performance over baselines, as well as gains from subject-specific fine-tuning \cite{Thuwajit2022EEGWaveNet:}. More recent seizure detection systems feature combinations of CNN and LSTM-style temporal modeling \cite{Poorani2023Deep}, an approach characteristic of seizure detection pipelines that learn spatiotemporal features to make binary or fixed-label predictions.

TUSZ is especially important because it is a large public benchmark with manual seizure annotations and eight seizure event types \cite{Amiri2025Epileptic, jiang2024large}. Those types include focal non-specific seizure (FNSZ) and generalized non-specific seizure (GNSZ), as well as more specific categories such as CPSZ, ABSZ, TNSZ, TCSZ, and MYSZ. Existing methods built on TUSZ and CHB-MIT generally assume that the label space observed during training is the same label space available at test time, whether the task is binary detection, seizure prediction, or multi-class seizure categorization \cite{Esmaeilpour2024Deep}. Even transfer-oriented seizure systems still fine-tune to a predefined set of downstream labels rather than modeling unknown seizure categories \cite{bakhtiarvand2026efm}. The central limitation, then, is that prior seizure-type classifiers are evaluated as fixed, closed-set predictors, rather than as systems that must remain reliable when rare or unseen seizure types arise in deployment.
\subsection{Open-Set Recognition}
Open-set recognition (OSR) considers a different deployment assumption: that the model should classify examples from known classes, but reject examples from unseen classes as unknown \cite{Bhatti2025Open}. Early OSR work formalizes open-set recognition as a tradeoff between empirical risk minimization on known classes and open-set risk minimization on unknown data, and this formulation largely motivates subsequent methods \cite{Chen2021Adversarial}. Methods considered since include discriminative, generative, and hybrid approaches such as OpenMax; reconstruction-based methods like C2AE or CROSR; and margin- or feature-space methods like ARPL \cite{Salehi2021A}. Now, post-hoc scoring rules such as MSP and its variants are considered common baselines.

Recent comparative studies suggest substantial overlap between OOD detection and OSR practice, and they show that simple scoring rules remain competitive, though performance can change markedly at larger scale \cite{Wang2024Dissecting}. In controlled OSR evaluations, Energy appears more stable than MSP across backbones, while feature-magnitude-sensitive rules often remain strong baselines \cite{Bhatti2025Open}. At the same time, better closed-set accuracy tends to correlate with better open-set performance, which makes strong pretrained backbones especially relevant for OSR \cite{Salehi2021A}. However, the modern OSR literature and the methods we compare against are evaluated primarily on vision benchmarks such as CIFAR and ImageNet-style settings. That leaves an important gap between OSR methodology and clinical EEG use, where unknowns arise from physiologically ambiguous or taxonomically rare seizure presentations rather than natural-image class holdouts \cite{Amiri2025Epileptic}.
\subsection{Clinical Motivation and Taxonomy
}
The justification for open-set seizure classification derives directly from the rationale behind the TUSZ seizure taxonomy. By defining not only specific seizure labels but also explicitly non-specific categories (FNSZ and GNSZ \cite{shah2018temple}), TUSZ accommodates an unavoidable clinical reality already embodied in the benchmark itself: seizures cannot always be unambiguously grouped into narrowly circumscribed types based on EEG data alone \cite{jiang2024large}. Beyond seizure-specific arguments, recent work on OSR benchmarking has made the case that semantic open-set evaluation is most meaningful when the known-unknown split is motivated by an explicit taxonomy rather than by arbitrary class partitioning \cite{Wang2024Dissecting}
This formulation lends OSR particularly well to seizure-typing use cases. If a model is presented with a seizure phenotype during inference that is not among the labels used during fine-tuning, we would expect the model to withhold a prediction or mark it as unknown rather than make an overconfident prediction by choosing the closest matching known class. To our knowledge, prior work in seizure detection or EEG foundation-model research does not pose the problem in this way; instead, they report performance on transfer, subject generalization, or closed-set classification with fixed downstream labels. Therefore, our setting differs in that we leverage the clinical seizure taxonomy itself to determine what should and should not be known at test time, connecting open-set machine learning to a clinically interpretable definition of seizure ambiguity and rarity.

Taken together, previous research has pushed the fields of EEG pretraining, seizure detection, and OSR along mostly separate lines of progress. What has been missing is a clinically motivated, open-set seizure-typing framework that leverages pretrained EEG representations and is tested on seizure types absent from fine-tuning.

\section{Methods}
\subsection{Problem Formulation}
We study open-set seizure type classification, where the label space is partitioned into a set of \emph{known} seizure classes.
\[
\mathcal{K}=\{\texttt{fnsz},\texttt{gnsz},\texttt{cpsz},\texttt{tcsz}\}
\]
and a disjoint set of \emph{unknown} seizure classes
\[
\mathcal{U}=\{\texttt{absz},\texttt{tnsz},\texttt{spsz},\texttt{mysz}\}.
\]
Given an EEG window $x$ with label $y \in \mathcal{K}\cup\mathcal{U}$, the model must correctly classify $x$ when $y\in\mathcal{K}$, and reject $x$ as unknown when $y\in\mathcal{U}$. Formally, let $f_\theta(x)\in\mathbb{R}^{|\mathcal{K}|}$ denote the closed-set logits over known classes only, and let $s(x)$ denote an open-set score derived from $f_\theta(x)$ and/or the penultimate feature representation. At inference time, the system predicts
\[
\hat{y}(x)=
\begin{cases}
\arg\max_{k\in\mathcal{K}} f_\theta(x)_k, & s(x)\ge \tau \\
\texttt{unknown}, & s(x)<\tau,
\end{cases}
\]
where $\tau$ is a rejection threshold calibrated on known-class windows from held-out validation subjects (Sec.~\ref{sec:score}). We define $\mathcal{K}$ using a practical subject-count threshold of at least 12 subjects per class, so that known classes support stable fine-tuning, while rarer seizure types are reserved for unknown evaluation.
\subsection{Dataset and Protocol}
\label{sec:dataset}
Experiments use TUSZ v2.0.3 and operate on 11{,}296 seizure windows from 206 subjects. We derive a 23-channel bipolar montage from the original 41-channel LE recordings; channels without a valid source in the target montage (FT9 and FT10) are zero-filled to preserve a fixed input shape. Each recording is segmented into non-overlapping 15\,s windows at 256\, Hz, yielding raw inputs of length 3840 samples per channel. Window labels are assigned by majority temporal overlap with the annotated seizure interval. To prevent label leakage across the open-set split, we treat subjects whose recordings contain both known and unknown seizure types as contaminated and exclude them from evaluation; this removes 4 subjects. The final protocol uses the class partition where \texttt{fnsz}, \texttt{gnsz}, \texttt{cpsz}, and \texttt{tcsz} serve as known classes and \texttt{absz}, \texttt{tnsz}, \texttt{spsz}, and \texttt{mysz} serve as true unknown classes. 

\paragraph{Subject-disjoint partition.}
For each random seed, we split the subjects of the known classes into disjoint fine-tuning (60\%), validation (20\%), and test (20\%) sets, stratified by class and processing the rarest class first; every subject belongs to exactly one set. Known-class test windows therefore always come from subjects \emph{not} seen during fine-tuning or validation. Table~\ref{tab:partition} reports subject and window counts per class. Because a few subjects contribute most windows of the rare classes, the
cpsz and tcsz test sets are small (3--5 subjects, 14--143 windows) and vary strongly across seeds; results involving these classes should be read with this in mind.

\begin{table}[t]
\centering
\caption{Known-class partition (mean over 3 seeds). Subjects are disjoint
across fine-tuning, validation, and test. Window counts vary across seeds,
most for the rare classes (e.g.\ tcsz test: 19--35 windows, cpsz test:
14--143).}
\label{tab:partition}
\small
\setlength{\tabcolsep}{5pt}
\begin{tabular}{lcccccc}
\toprule
 & \multicolumn{2}{c}{Fine-tune} & \multicolumn{2}{c}{Val} & \multicolumn{2}{c}{Test}\\
\cmidrule(lr){2-3}\cmidrule(lr){4-5}\cmidrule(lr){6-7}
Class & Subj & Win & Subj & Win & Subj & Win\\
\midrule
fnsz & 52 & 2400 & 17 & 1005 & 18 & 747\\
gnsz & 69 & 3072 & 20 & 1061 & 20 & 1181\\
cpsz & 15 & 532  & 5  & 104  & 5  & 88\\
tcsz & 7  & 151  & 3  & 116  & 3  & 28\\
\bottomrule
\end{tabular}
\end{table}
\subsection{EEG Foundation Model}
The backbone is a pretrained EEG foundation model with approximately 17M parameters. It contains a raw-signal encoder based on Mamba, a time-frequency encoder based on a Transformer, and a fusion MLP that combines both streams into a shared representation. For each 15\,s EEG window, the model takes as input a raw tensor of shape $(23,3840)$ and a continuous-wavelet (CWT) scalogram of shape $(23,64,3840)$. The scalogram is computed with a Morlet wavelet over 0.5--70\, Hz using 64 frequency bins at a sampling rate of 256\, Hz. The fused output is a 256-dimensional embedding $\mathbf{e}_{\mathrm{fused}} \in \mathbb{R}^{256}$, which is passed to a lightweight classification head during fine-tuning. Large-scale pretrained classifiers are a natural backbone for post-hoc open-set scoring because both energy-based and feature-based OSR methods are explicitly designed to operate on pre-trained networks without changing the base architecture. 

\paragraph{Pretraining data and overlap with TUSZ.}
The EFM was pretrained without labels on 10{,}000 EDF recordings from the TUH EEG corpus (TUEG) and on SEED~\cite{bakhtiarvand2026efm}. TUSZ was not
explicitly excluded. Because TUSZ is drawn from TUEG, 30 of our 206 subjects (30 of 350 sessions) also appear in the pretraining data, accounting for at most 12\% of known-class windows. Of the 17 subjects with true-unknown seizure types, one (absz) appears in pretraining, and none of their seizure sessions do. Pretraining used no seizure annotations, so no
seizure-type labels leaked into the backbone, but the encoder has seen unlabeled EEG from a minority of our subjects.
\subsection{Fine-Tuning Strategy}
Fine-tuning is parameter-efficient. We freeze the adapters, patch projections, Mamba blocks 0-3, and time-frequency Transformer blocks 0-3. We update Mamba blocks 4-5, the top two time-frequency Transformer layers, the fusion MLP, and a newly initialized classifier head, corresponding to 17.3\% of model parameters. The head is
\[
\mathbf{h} = \mathrm{Dropout}(\mathrm{GELU}(\mathbf{W}_1 \mathbf{e}_\text{fused} + \mathbf{b}_1)),
\quad
\mathbf{f}_\theta(x) = \mathbf{W}_2 \mathbf{h} + \mathbf{b}_2,
\]

where $\mathbf{W}_1 \in \mathbb{R}^{128 \times 256}$ and $\mathbf{W}_2 \in \mathbb{R}^{K \times 128}$.
Training minimizes weighted cross-entropy with inverse-frequency class weights using AdamW and a cosine learning-rate schedule for up to 50 epochs, with early stopping based on validation accuracy and patience 7. This produces a closed-set classifier over known classes, after which open-set behavior is induced entirely through post-hoc scoring.

\subsection{Open-Set Scorers}
\label{sec:score}
\paragraph{Maximum softmax probability (MSP).}
As an additional baseline, we threshold the maximum softmax probability ~\cite{Hendrycks2017baseline}: $s(x) = \max_k \, \mathrm{softmax}(f_\theta(x))_k$.

\paragraph{Energy score.}
We use the energy score~\cite{Liu2020Energy}:
\[
E(x) = -T \log \sum_{k=1}^{K} \exp\left(\frac{f_\theta(x)_k}{T}\right),
\]
with temperature $T=1$ unless otherwise stated. Energy operates directly in logit space rather than softmax space, which preserves information lost by the normalization implicit in softmax confidence. It applies to any pre-trained classifier without retraining and is parameter-free at inference time. We treat larger in-distribution evidence as more likely known, and classify a sample as unknown when its calibrated energy falls below threshold $\tau$.

\paragraph{GHOST.}
GHOST~\cite{ghost2025} models penultimate-layer features with a class-wise multivariate Gaussian distribution with diagonal covariance for each known class. Let $\mathbf{e}(x)$ denote the fused embedding and let $(\boldsymbol{\mu}_k,\boldsymbol{\sigma}_k^2)$ be the mean and diagonal variance estimated from training embeddings of class $k$. We compute a per-class Gaussian compatibility score:
\[
g_k(x) = -\frac{1}{2}\sum_{d}\left[\frac{(e_d(x)-\mu_{k,d})^2}{\sigma_{k,d}^2} + \log \sigma_{k,d}^2\right]
\]
and combine it with a z-score normalization of logits to reduce the effect of abnormal feature magnitudes that can otherwise give unknown samples spuriously high class scores. GHOST is hyperparameter-free and was introduced specifically to improve both overall OSR performance and fairness across classes, since aggregate metrics can hide large class-to-class disparities.

\paragraph{COSTARR.}
COSTARR~\cite{costarr2025} is motivated by the attenuation hypothesis, which states that small classifier weights suppress feature dimensions that are uninformative for a given known class but may still contain useful evidence for distinguishing known from unknown inputs. For an input feature vector $\mathbf{F}$ and class weight vector $\mathbf{W}_j$, COSTARR forms a post-attenuation representation using the Hadamard product
\[
\mathbf{H}_j = \mathbf{F} \odot \mathbf{W}_j,
\]
then combines pre-attenuation features, post-attenuation features, and the logit of the predicted class into a consolidated score. This design captures unexpected activation patterns in attenuated dimensions that standard confidence scores ignore. COSTARR is post-hoc, architecture-agnostic, and adds minimal computation at test time.

\paragraph{Threshold calibration.}
For each scorer, we set the rejection threshold $\tau$ to the 5th percentile of its scores on known-class windows from the
\emph{validation} subjects (Sec.~3.2), targeting a 5\% false rejection rate of known windows. Validation subjects are disjoint from both the fine-tuning and the test subjects, so $\tau$ is never calibrated on data the classifier was trained on or evaluated on. All known-class scores we report are computed on the held-out test subjects. A fixed threshold
converts the continuous familiarity score into a known-versus-unknown decision, matching the practical role of score-based OSR.

\subsection{LOCO Evaluation Protocol}
We evaluate open-set performance with four leave-one-class-out (LOCO) rounds. In each round, one class from $\mathcal{K}$ is removed from fine-tuning and treated as a \emph{LOCO unknown}; the model is then fine-tuned from scratch on the remaining three known classes, so $K=3$ for that round. Because many subjects have more than one seizure type, some windows of the
held-out class come from subjects whose \emph{other} seizure types were used for fine-tuning in that round. Subject identity is therefore a potential confound: a scorer may reject held-out windows partly because the \emph{subject} is new rather than because the seizure \emph{type} is new. We therefore split the held-out-class windows into those from subjects \emph{unseen} during fine-tuning, which form our primary LOCO
evaluation set, and those from subjects \emph{seen} during fine-tuning, which we report separately as a control (Sec.~\ref{sec:fairness}). Known-class windows are always taken from the held-out test subjects. At test time, we evaluate two forms of novelty: (i) the held-out known class, which measures whether the method can reject a clinically relevant but unseen seizure type close to the training taxonomy, and (ii) the true unknown set $\mathcal{U}=\{\texttt{absz},\texttt{tnsz},\texttt{spsz},\texttt{mysz}\}$, which measures detection of rare seizure types absent from both fine-tuning and the known-class pool. Subjects with true-unknown seizures never appear in fine-tuning, so the
same subject-novelty caveat applies to this set. We report AUROC, AUOSCR, CCR at FPR $\in\{0.05, 0.10, 0.20\}$, F@C95,
and the coefficient of variation (COV) of per-subject rejection rates as a fairness metric, each as mean $\pm$ standard deviation over three seeds.

\subsection{Clinical relevance of CCR@FPR operating points.}
We report CCR at three FPR thresholds: 5\%, 10\%, and 20\%.
These correspond to clinically meaningful trade-offs between
sensitivity and specificity in a deployed alarm system. At
FPR\,=\,5\%, the system accepts only the most confidently
known windows, minimizing false alarms at the cost of rejecting more true known-class events. At FPR\,=\,20\%, the system is more permissive, accepting a broader range of inputs and recovering more correct classifications at the expense of occasionally accepting unknown-type windows. In clinical practice, the appropriate operating point depends on the deployment context: a screening system tolerates higher FPR to avoid missing events, while a system that triggers interventions requires stricter control. Reporting across all three thresholds allows clinicians and system designers to select the operating point that matches their safety requirements rather than committing to a single arbitrary threshold.

\section{Experiments and Results}
\subsection{Implementation Details}

All experiments run on a server with 8 $ \ times$ NVIDIA V100 16\, GB GPUs.
Fine-tuning uses the AdamW optimizer~\cite{Loshchilov2019AdamW} with
learning rate $1\times10^{-4}$ and weight decay $10^{-4}$.
The learning rate follows a cosine annealing schedule with
$\eta_\text{min} = 5\times10^{-5}$, a maximum of 50 epochs, and early stopping with patience 7, monitored on validation accuracy over the validation subjects (Sec.~\ref{sec:dataset}). We use mini-batches of size 16 throughout.
CWT scalograms are computed on-the-fly per batch using the Morlet wavelet with 64 log-linearly spaced over $0.5$--$70$\, Hz at
$f_s=256$\, Hz. No scalograms are cached to disk.
Class imbalance is addressed with inverse-frequency weights in the
cross-entropy loss:
\[
w_k = \frac{N}{K \cdot N_k},
\]
where $N$ is the total number of training windows, $K$ the number of
known classes, and $N_k$ the window count for class $k$.
For each LOCO round, a fresh model is initialized from the pretrained
EFM checkpoint and fine-tuned independently; no information from the
held-out class leaks into any round.

\paragraph{Reproducibility.}
We repeat every experiment with three seeds (42, 43, 44). Each seed defines its own subject partition, fine-tuning run, and scorer fit, and we report mean $\pm$ standard deviation across seeds. Code, the exact subject splits, and the scripts that generate all tables and figures are publicly available\footnote{\url{https://github.com/Targolb/openset-seizure-tusz}};
the pretrained EFM checkpoint is available for research use on request.
\subsection{Closed-Set Classification}
Before evaluating open-set detection, we report how well the fine-tuned model classifies the three training classes in each LOCO round.

Closed-set accuracy on the held-out test subjects ranges from 0.55 to 0.74 across rounds (Table~\ref{tab:closed}) and varies by up to 0.11 across seeds, reflecting how strongly a few subjects shape each partition. The classifier is therefore a moderate, not strong, closed-set model, which limits how well any post-hoc scorer can separate known from unknown.

\begin{table}[t]
\centering
\caption{Closed-set accuracy on known-class test subjects (mean $\pm$ std over seeds).}
\label{tab:closed}
\small
\begin{tabular}{lc}
\toprule
Held-out class & Accuracy\\
\midrule
fnsz & 0.74 $\pm$ 0.07\\
gnsz & 0.72 $\pm$ 0.11\\
cpsz & 0.58 $\pm$ 0.10\\
tcsz & 0.55 $\pm$ 0.09\\
\bottomrule
\end{tabular}
\end{table}
\subsection{Open-Set Detection: True Unknowns}

Table~\ref{tab:openset} reports results for seven post-hoc scorers. On true unknowns, Mahalanobis has the highest mean on every metric (AUROC 0.662, AUOSCR 0.470, CCR@10 0.261), followed by kNN and Energy; GHOST and COSTARR are below Energy, and MSP is weakest. The gaps between the top scorers (about 0.02-0.03 AUROC) are comparable to the seed-to-seed standard deviation, so we do not claim a reliable ranking. Performance also varies strongly by round: true-unknown AUROC is highest when gnsz is held out (up to 0.78) and near chance when tcsz is held out (0.48–0.55).

\begin{figure}[t]
\centering
\includegraphics[width=\columnwidth]{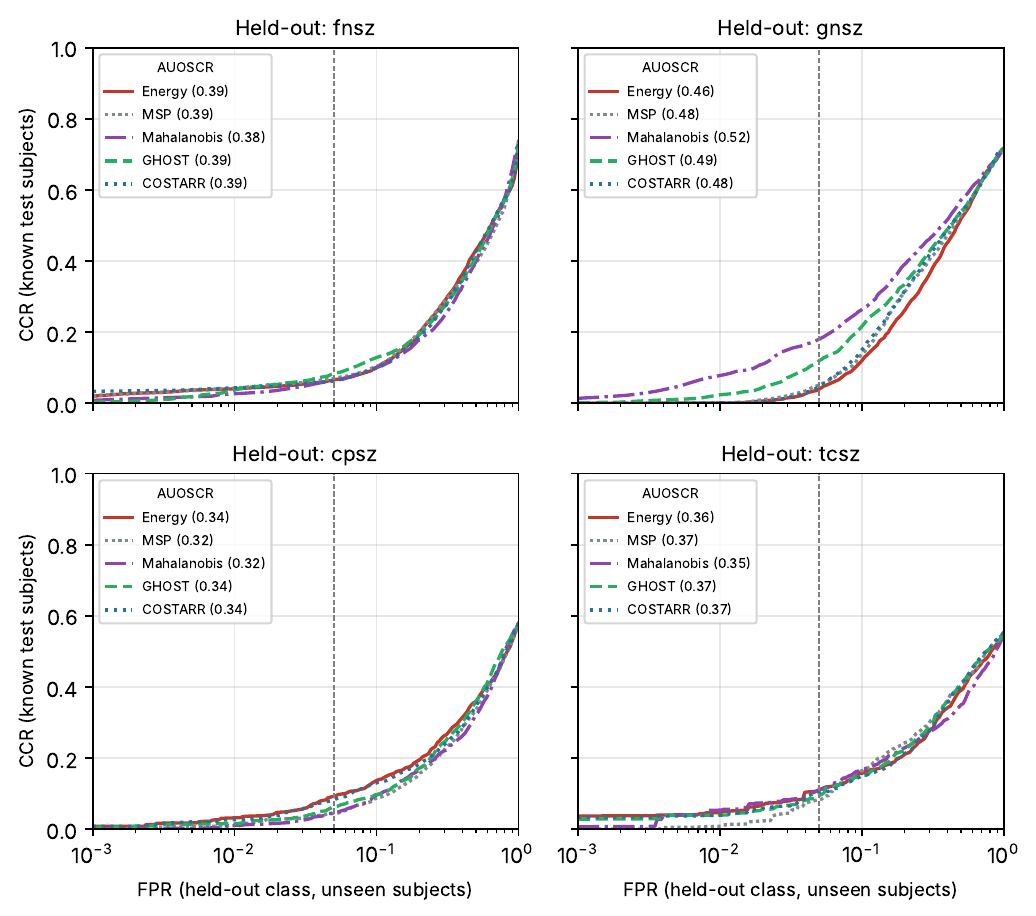}
\caption{LOCO OSCR curves (log-scale FPR): correct classification rate on
known-class test subjects vs.\ false positive rate on held-out-class windows
from unseen subjects, averaged over three seeds. Dashed line:
FPR = 0.05. Legend values are mean AUOSCR. Curves overlap in all rounds
except gnsz, where Mahalanobis and GHOST are higher at low FPR.}
\label{fig:oscr}
\end{figure}

On LOCO unknowns from unseen subjects, all scorers are close to chance (AUROC 0.54--0.58, CCR@10 $\le$ 0.16), and their differences are within one standard deviation. Only the gnsz round is consistently above chance (Fig.~\ref{fig:oscr}). Across scorers, true unknowns are easier to detect
than LOCO unknowns (mean AUROC 0.62 vs.\ 0.57).

\begin{table}[t]
\centering
\caption{Open-set results averaged over the four LOCO rounds (mean $\pm$
std over seeds). Top: true unknowns $\mathcal{U}$. Bottom: held-out class,
unseen subjects. $\uparrow$ higher is better; F@C95: FPR at 95\% of
closed-set accuracy, $\downarrow$ lower is better.}
\label{tab:openset}
\resizebox{\columnwidth}{!}{%
\begin{tabular}{lcccccc}
\toprule
Scorer & AUROC$\uparrow$ & AUOSCR$\uparrow$ & CCR@5$\uparrow$ & CCR@10$\uparrow$ & CCR@20$\uparrow$ & F@C95$\downarrow$\\
\midrule
\multicolumn{7}{l}{\emph{True unknowns}}\\
Energy      & .642$\pm$.050 & .436$\pm$.031 & .175$\pm$.015 & .224$\pm$.015 & .305$\pm$.026 & .850$\pm$.050\\
MSP         & .551$\pm$.062 & .382$\pm$.037 & .084$\pm$.030 & .126$\pm$.030 & .228$\pm$.054 & .874$\pm$.041\\
MaxLogit    & .626$\pm$.048 & .429$\pm$.030 & .157$\pm$.027 & .214$\pm$.020 & .300$\pm$.025 & .846$\pm$.040\\
Mahalanobis & .662$\pm$.028 & .470$\pm$.020 & .186$\pm$.024 & .261$\pm$.018 & .350$\pm$.013 & .764$\pm$.091\\
kNN         & .634$\pm$.015 & .447$\pm$.010 & .164$\pm$.015 & .223$\pm$.010 & .316$\pm$.015 & .819$\pm$.027\\
GHOST       & .594$\pm$.021 & .419$\pm$.016 & .141$\pm$.014 & .193$\pm$.005 & .278$\pm$.003 & .841$\pm$.052\\
COSTARR     & .612$\pm$.043 & .423$\pm$.028 & .151$\pm$.025 & .206$\pm$.014 & .291$\pm$.016 & .866$\pm$.042\\
\midrule
\multicolumn{7}{l}{\emph{LOCO (held-out class, unseen subjects)}}\\
Energy      & .583$\pm$.024 & .390$\pm$.012 & .077$\pm$.028 & .129$\pm$.031 & .216$\pm$.029 & .912$\pm$.022\\
MSP         & .568$\pm$.025 & .389$\pm$.008 & .062$\pm$.010 & .124$\pm$.006 & .229$\pm$.011 & .902$\pm$.012\\
MaxLogit    & .581$\pm$.025 & .392$\pm$.013 & .077$\pm$.028 & .129$\pm$.030 & .216$\pm$.028 & .904$\pm$.024\\
Mahalanobis & .544$\pm$.024 & .393$\pm$.023 & .100$\pm$.027 & .157$\pm$.035 & .241$\pm$.045 & .898$\pm$.027\\
kNN         & .561$\pm$.024 & .398$\pm$.016 & .098$\pm$.019 & .161$\pm$.015 & .244$\pm$.028 & .892$\pm$.035\\
GHOST       & .575$\pm$.018 & .400$\pm$.012 & .090$\pm$.012 & .149$\pm$.013 & .231$\pm$.013 & .894$\pm$.031\\
COSTARR     & .582$\pm$.026 & .395$\pm$.013 & .074$\pm$.028 & .132$\pm$.023 & .225$\pm$.023 & .883$\pm$.038\\
\bottomrule
\end{tabular}}
\end{table}

\subsection{Fairness Analysis}
\label{sec:fairness}
We compute, for each subject with at least five unseen-subject LOCO windows,
the fraction of its windows rejected at the threshold that rejects 10\% of
known test windows, and summarize spread with the coefficient of variation
(Table~\ref{tab:cov}). COV values are high for all scorers (0.7--1.8):
most subjects have rejection rates near zero and a few are rejected often.
GHOST has a lower mean COV than Energy for fnsz, cpsz, and tcsz, but with
large seed-to-seed variation. To test the difference, we average each
subject's rejection rate over the seeds in which it is unseen and apply a
two-sided Wilcoxon signed-rank test (GHOST vs.\ Energy, paired by subject);
no round shows a significant difference ($p \ge 0.25$). We therefore find
no evidence that GHOST improves per-subject fairness in this setting.
\begin{table}[t]
\centering
\caption{COV of per-subject rejection rate on LOCO unknowns from unseen
subjects ($\ge$5 windows), threshold at 10\% known false rejection (mean
$\pm$ std over seeds). Lower = more uniform. $p$: two-sided Wilcoxon
signed-rank test on per-subject rejection rates, GHOST vs.\ Energy.}
\label{tab:cov}
\small
\setlength{\tabcolsep}{4pt}
\begin{tabular}{lcccc}
\toprule
Held-out & $n$ subj & Energy COV & GHOST COV & $p$\\
\midrule
fnsz & 57 & 1.77 $\pm$ 0.50 & 1.59 $\pm$ 0.24 & 0.94\\
gnsz & 77 & 1.08 $\pm$ 0.17 & 1.08 $\pm$ 0.22 & 0.25\\
cpsz & 11 & 1.42 $\pm$ 0.49 & 1.14 $\pm$ 0.46 & 0.37\\
tcsz & 8 & 0.97 $\pm$ 0.17 & 0.74 $\pm$ 0.04 & 0.81\\
\bottomrule
\end{tabular}
\end{table}

\subsection{Subject Confound in LOCO}
\label{sec:subject_confound}

Table~\ref{tab:seen} lists, for each held-out class, how many of its windows come from subjects whose other seizure types were used for fine-tuning. For fnsz and gnsz this is a substantial fraction (about 15\% and 20--30\% of windows, from 23--27 subjects); for cpsz and tcsz only 2--5 subjects are seen, so the comparison is not informative for those
rounds. If the scorers mainly detected subject novelty, AUROC would be higher on unseen-subject than on seen-subject windows. We observe this for gnsz (Energy $0.60$ vs.\ $0.51$; GHOST $0.63$ vs.\ $0.55$), but not for
fnsz, where both subsets give similar, near-chance AUROC. Part of the gnsz signal may therefore reflect new subjects rather than a new seizure type, and we cannot rule out the same effect in the other rounds or for true unknowns, whose subjects never appear in fine-tuning.

\begin{table}[t]
\centering
\caption{LOCO held-out-class windows from subjects seen during fine-tuning,
and AUROC (Energy / GHOST) on unseen vs.\ seen subjects, mean over seeds.}
\label{tab:seen}
\small
\setlength{\tabcolsep}{3pt}
\begin{tabular}{lccccc}
\toprule
 & \multicolumn{3}{c}{Windows} & \multicolumn{2}{c}{AUROC (En / GH)}\\
\cmidrule(lr){2-4}\cmidrule(lr){5-6}
Class & All & Seen & Seen subj & Unseen & Seen\\
\midrule
fnsz & 4152 & 636  & 24 & .53 / .50 & .53 / .53\\
gnsz & 5314 & 1280 & 27 & .60 / .63 & .51 / .55\\
cpsz & 724  & 291  & 5  & .56 / .53 & .63 / .61\\
tcsz & 295 & 72 & 3 & .64 / .63 & .58 / .62\\
\bottomrule
\end{tabular}

\end{table}

\subsection{Discussion}
\paragraph{LOCO vs.\ true unknowns.}
Averaged over scorers, true unknowns are easier to detect than LOCO unknowns (mean AUROC 0.62 vs.\ 0.57), but the pattern is not uniform: in the tcsz round, true unknowns are detected near chance (0.48--0.55) while held-out tcsz windows are detected above it. One possible explanation is that true-unknown types are more distant from the training classes in embedding space, whereas fnsz and gnsz are defined by the absence of distinguishing features and overlap with the remaining known types. However, true-unknown subjects never appear in fine-tuning, so part of this gap may reflect subject novelty rather than seizure-type novelty (Sec.~\ref{sec:subject_confound}); our data cannot separate the two.
\paragraph{Limitations.}
This benchmark should be read as an initial baseline. First, cohorts are small for the rare known classes: cpsz and tcsz test sets contain only 3--5 subjects, and per-subject fairness estimates rely on as few as 8 subjects, so results for these rounds are noisy. Second, although known-class test subjects are disjoint from fine-tuning subjects, LOCO unknowns can come from subjects seen during fine-tuning, and true unknowns
come from subjects never seen at all; the scorers may therefore partly detect new \emph{subjects} rather than new seizure \emph{types}, and our seen/unseen split (Sec.~\ref{sec:subject_confound}) bounds this effect only for fnsz and gnsz. Third, the EFM was pretrained on TUEG recordings
that include 30 of our subjects (no seizure labels. One true unknown subject, none of its seizure sessions). Fourth, results depend strongly on the subject partition (closed-set accuracy 0.43--0.86 across seeds and rounds), and three seeds give only rough uncertainty estimates. Fifth, all metrics are computed at the window level; event- or patient-level evaluation, which is closer to clinical use, is left for future work.
Finally, we evaluate a single EFM checkpoint and only post-hoc scorers. Training-time open-set methods and other backbones may behave differently.


\section{Conclusion}

We introduced the first open-set recognition protocol for EEG seizure
type classification on TUSZ v2.0.3, addressing a critical gap between
how automated seizure classifiers are evaluated and how they behave in
clinical deployment. Our protocol defines known and unknown seizure
classes using the clinical taxonomy itself, and evaluates seven post-hoc OSR scorers applied to a partially fine-tuned EEG Foundation Model.

Under a subject-disjoint protocol with three seeds, we find that (i) post-hoc scorers detect true unknown seizure types only moderately (best mean AUROC 0.66, Mahalanobis), with no reliable ranking among the top scorers; (ii) held-out known seizure types from unseen subjects are detected only slightly above chance by all scorers (AUROC 0.54–0.58); and (iii) per-subject rejection rates are highly uneven, with no significant fairness difference between GHOST and energy scoring.

We hope this protocol and benchmark serve as a foundation for future
work on clinically deployable seizure typing systems that recognize the
limits of their label space rather than forcing incorrect predictions.
Future directions include few-shot adaptation for rare seizure types,
cross-dataset evaluation, and extending the fairness analysis to
demographic subgroups.

This work represents a first step towards clinically deployable open-set seizure type recognition. Seizure type classification under the open-set assumption is a genuinely hard problem: the known classes are defined partly by the absence of distinguishing features (\texttt{fnsz}, \texttt{gnsz}), the unknown classes are rare by definition, and the available data is severely imbalanced across both subjects and types.Our results show that, with a general-purpose EFM and lightweight fine-tuning, current post-hoc scorers detect novel seizure types only weakly, and substantial work remains. Closing the gap between benchmark performance and clinical deployment will require models pretrained with explicit open-set objectives, larger and more diverse seizure corpora, and prospective validation on real clinical workflows where seizure type labels are uncertain, delayed, or absent entirely.

{
    \small
    \bibliographystyle{ieeenat_fullname}
    \bibliography{main}
}

\end{document}